\documentclass[runningheads]{llncs}

\usepackage{eccv}

\usepackage{eccvabbrv}

\usepackage{graphicx}
\usepackage{booktabs}
\usepackage{multirow}
\usepackage{amsmath}
\usepackage{inconsolata}
\usepackage{algorithm}
\usepackage{algpseudocode}
\usepackage{graphicx}
\usepackage{subcaption}
\usepackage{bbm}

\usepackage[accsupp]{axessibility}  

\usepackage{hyperref}

\usepackage{orcidlink}

\begin{document}

\title{Spectral Optimal Transport for Dynamics-Aware Sign Language Generation} 

\titlerunning{SOT-SLG}

\author{Nabeela Khan\inst{1,2}\orcidlink{0009-0007-7105-9076} \and
Bowen Wu\inst{2,3}\orcidlink{0000-0001-8518-2939} \and
Runwu Shi\inst{1}\orcidlink{0000-0001-9097-9619} \and
Hanlong Li\inst{1}\orcidlink{0009-0002-2579-2107} \and
Benjamin Yen\inst{4,1}\orcidlink{0000-0002-6958-7319} \and
Carlos Toshinori Ishi\inst{2,3}\orcidlink{0009-0003-1225-4702} \and
Kazuhiro Nakadai\inst{1}\orcidlink{0000-0002-6134-4558}}

\authorrunning{N.~Khan et al.}

\institute{Institute of Science Tokyo, Tokyo, Japan\\
\email{\{nabeela,shirunwu,lihanlong,nakadai\}@ra.sc.eng.isct.ac.jp} \and
RIKEN Guardian Robot Project, Kyoto, Japan\\
\email{\{bowen.wu,carlos.ishi\}@riken.jp} \and
ATR Hiroshi Ishiguro Laboratories, Kyoto, Japan \and
RIKEN Center for Biosystems Dynamics Research, Kobe, Japan \\
\email{benjamin.yen@ieee.org}}

\maketitle

\begin{abstract}

Sign language generation (SLG) translates spoken-language text into pose sequences that convey its meaning. Recently, flow-based models with optimal transport (OT) have become a strong SLG baseline by handling the one-to-many nature of text-to-sign. While classifier-free guidance (CFG) could be applied to further strengthen the semantic alignment theoretically, its training involves conditional and unconditional training, which may contradict the pose-space OT, thus reducing the generation quality. Under the condition dropout, one noise sample can be paired with target motions exhibiting incompatible motion dynamics, interfering with the semantic guidance when the text condition is present. To address this issue, we propose Spectral Optimal Transport (SOT), which augments the pose-space cost with a spectral feature distance, reweighting the coupling toward the temporal-frequency structure of the target motions, thereby improving the performance when using CFG. The quantitative and qualitative experiments demonstrate the effectiveness of the proposed method. The code is available at \url{https://github.com/foxrei-n/SignFlow}.
  \keywords{Sign language generation \and Conditional flow matching \and Classifier-free guidance \and Motion dynamics modeling}

\end{abstract}

\section{Introduction}
\label{sec:intro}

\begin{figure*}[t]
  \centering

  \begin{minipage}[t]{0.48\textwidth}
    \centering
    \includegraphics[width=0.85\linewidth]{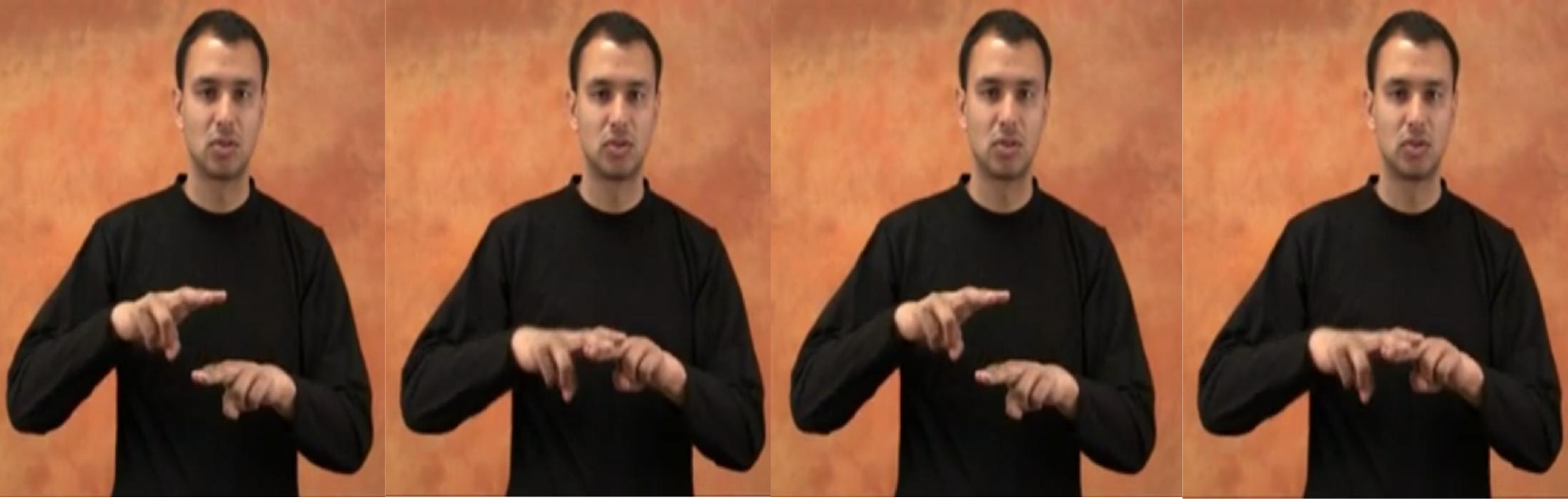}\\[-1mm]
    {\small \textit{``Chair''}}
  \end{minipage}\hfill
  \begin{minipage}[t]{0.48\textwidth}
    \centering
    \includegraphics[width=0.85\linewidth]{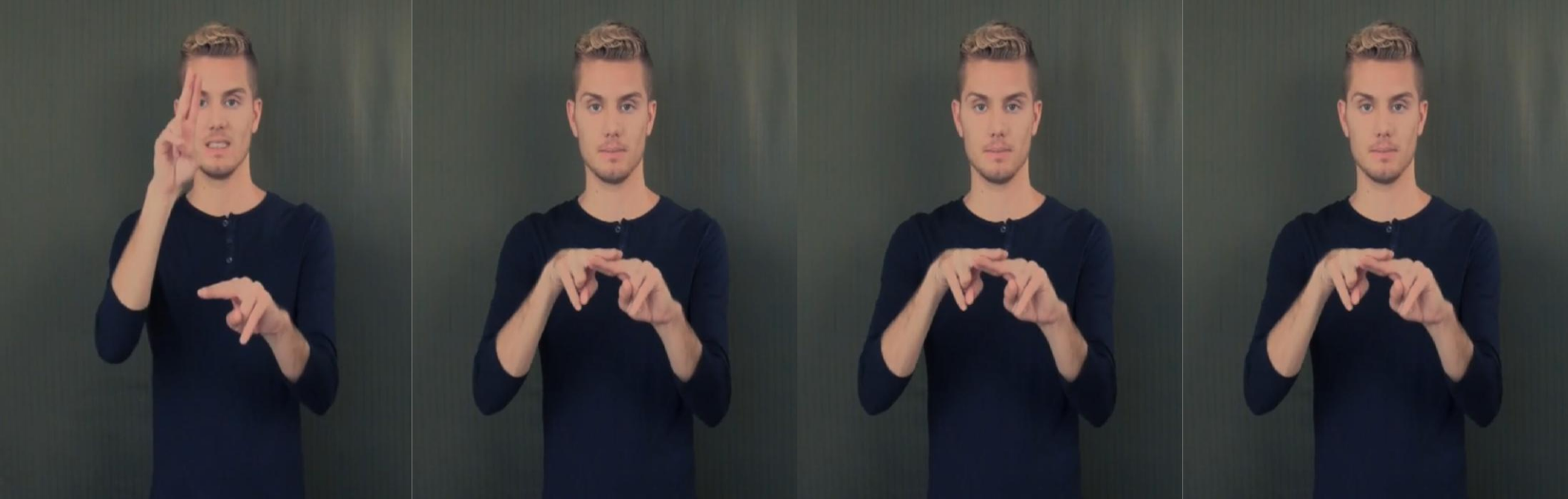}\\[-1mm]
    {\small \textit{``Sit''}}
  \end{minipage}

  {\small\bfseries (a) Repetition vs. single motion}
  \\[5mm]

  \begin{minipage}[t]{0.48\textwidth}
    \centering
    \includegraphics[width=0.85\linewidth]{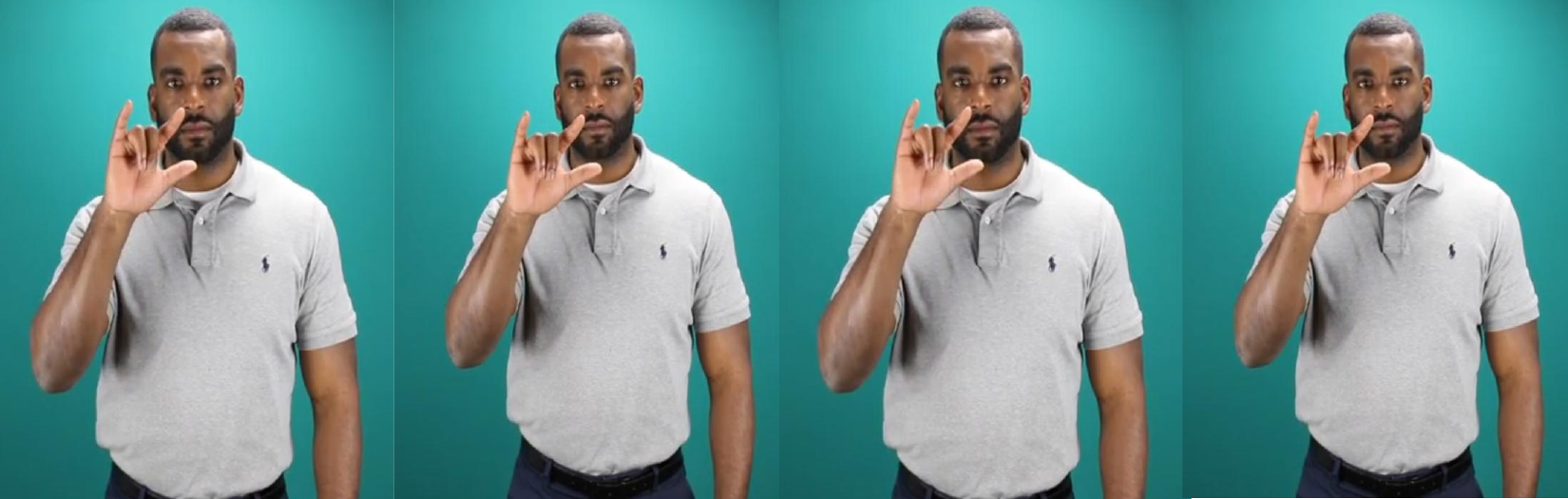}\\[-1mm]
    {\small \textit{``Airplane''}}
  \end{minipage}\hfill
  \begin{minipage}[t]{0.48\textwidth}
    \centering
    \includegraphics[width=0.85\linewidth]{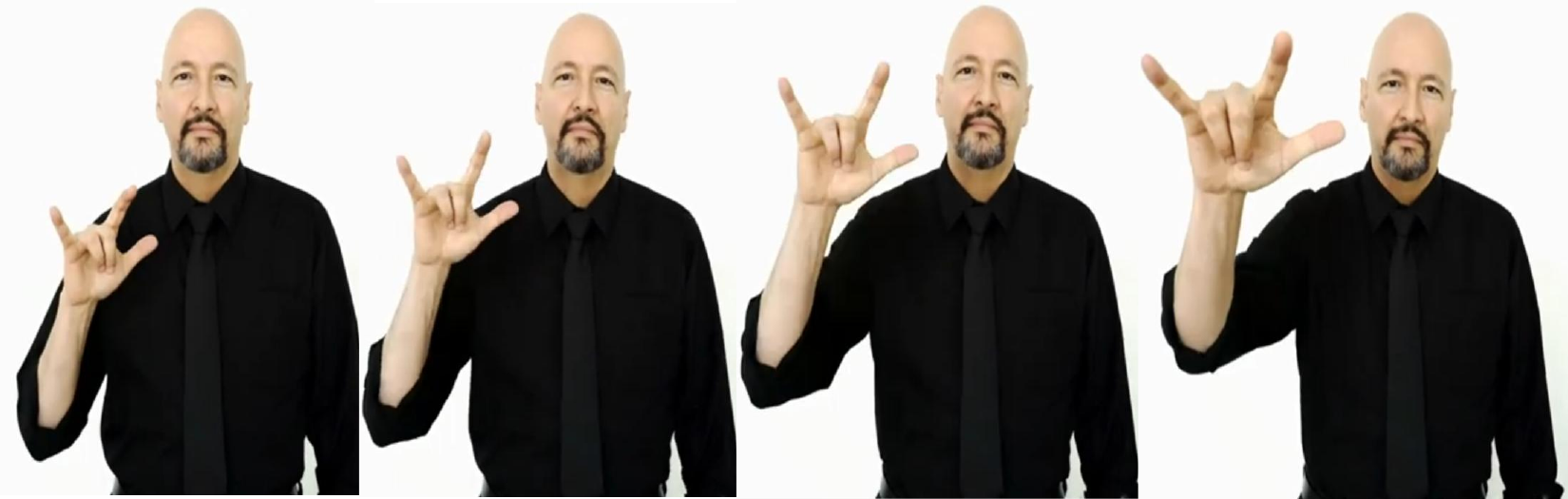}\\[-1mm]
    {\small \textit{``Fly''}}
  \end{minipage}

  {\small\bfseries (b) Small vs.\ large path extent}

  \caption{Minimal pairs illustrating the phonological role of motion dynamics in sign language. The images are sourced from WLASL \cite{li2020word}.}
  \label{fig:minimal_pairs}
\end{figure*}

Sign languages (SLs) are full-fledged natural languages used by Deaf and Hard-of-Hearing (DHH) communities \cite{yin2021including}. They are expressed in the visual--manual modality through coordinated manual and non-manual elements \cite{sandler2006sign}. Sign language generation (SLG) aims to generate sign pose sequences conditioned on spoken-language text, where a central goal is \emph{semantic alignment}, meaning the generated motion should convey the intended meaning. Prior work has improved SLG by enhancing motion realism \cite{baltatzis2024neural}, reducing prediction drift in long sequences \cite{khan2025multigau}, handling information density \cite{yin2024t2s}, and incorporating expressive factors such as emotions \cite{qi2024signgen}. Recently, SignFlow \cite{khan2025signflow} emerged as a strong flow-based SLG model that better handles the \emph{one-to-many} nature of SLG via conditional flow matching and minibatch optimal transport (CFM-OT). However, because the same text can correspond to multiple valid realizations, the text alone does not uniquely determine the pose sequence at inference. To improve adherence to the input text, we use the standard classifier-free guidance (CFG) procedure \cite{ho2022classifier}.
Empirically, we confirm that CFG improves semantic alignment; however, improved semantic alignment does not necessarily guarantee that the generated motion realizes the motion dynamics required for linguistically meaningful signing. Generally, we view such motion dynamics as part of aligned semantic realization. 
SLs exhibit a structured phonology \cite{stokoe2005sign,emmorey1995comparison} where signs are composed of sublexical features such as handshape, location, movement, and orientation \cite{emmorey1995comparison}, and meaning can depend critically on \emph{how motion evolves over time} (e.g., repetition vs.\ single movement, fast vs.\ slow articulation) \cite{bosworth2019analysis}. Figure~\ref{fig:minimal_pairs} illustrates minimal contrasts where the main difference is motion dynamics: \textit{chair} vs.\ \textit{sit}, differs by repeated vs.\ single movement, and \textit{airplane} vs.\ \textit{fly} differs by repeated small motions vs.\ a larger single motion. These cues are not merely stylistic; suppressing them can yield visually plausible but linguistically incorrect signs \cite{hwang2025efficient}. Importantly for SLG, such dynamics also vary across signers and contexts, so even identical text can correspond to multiple valid realizations with distinct motion dynamics.

In SignFlow, minibatch OT computes a permutation that pairs prior noise samples with target motion sequences, shaping the supervision for the learned vector field \cite{khan2025signflow}. CFM-OT\cite{tong2023improving} uses a Euclidean pose-space coupling cost that is largely insensitive to motion dynamics. This becomes problematic under CFG-style condition dropout, where the same network is trained on both unmasked (text-conditioned) and masked (unconditioned) examples. Conceptually, unmasked updates are driven by a conditional motion distribution that is relatively concentrated for a given sentence, while masked updates are driven by a broader marginal motion distribution that mixes diverse dynamics across the dataset. This mismatch makes the masked and unmasked training signals less compatible, which can limit the effectiveness of guidance at inference when the two branches are combined.

We therefore propose Spectral Optimal Transport (SOT), a frequency-aware OT coupling that augments the assignment cost with a spectral distance computed from temporal DCT (Discrete Cosine Transform) features, making the assignment sensitive to the temporal-frequency structure of the target motions. Prior work has modeled motion dynamics (trajectories) with DCT bases \cite{akhter2008nonrigid}, and later motion prediction methods used frequency-space formulations, often to promote smoothness \cite{mao2019learning,tian2024transfusion}.
In SLG, however, preserving motion dynamics (e.g., repetition and path extent) is important for faithful realizations, and these dynamics may vary across valid sequences for the same text.
Rather than comparing motion dynamics between the two operands, a prior sample has none; SOT reweights the metric under which noise and targets are coupled: mismatches in higher temporal-frequency components contribute more to the assignment cost, so targets that differ in motion dynamics become more distinguishable during coupling. This reduces masked--unmasked inconsistency under CFG training and amplifies the gains from guidance.
 
This paper makes the following contributions:
\begin{itemize}
    \item We empirically demonstrated that classifier-free guidance (CFG) improves semantic alignment for flow-based text-to-sign language generation.
    \item We proposed Spectral Optimal Transport (SOT), a simple frequency-aware extension of minibatch OT that reweights the coupling cost toward temporal-frequency structure, so that the assignment becomes sensitive to the motion dynamics of the targets.
    \item We showed that SOT reduces masked-unmasked mismatch and amplifies the gains from CFG.
\end{itemize}

\section{Related Work}

\subsection{Sign Language Generation}

A large number of SLG works rely on \textit{glosses}\footnote{A gloss is a label representing an isolated sign, often used as an intermediate symbolic representation.} as an intermediate representation, following a text$\rightarrow$gloss$\rightarrow$pose (or video) pipeline
\cite{huang2021towards,saunders2022signing,tang2022gloss,xie2024g2p,walsh2024select,tang2025gloss,zuo2024simple,xie2024sign,zhang2025towards}.
While glosses can simplify learning by providing a discrete gloss-token interface, they require costly expert annotation and introduce an intermediate bottleneck that may discard linguistically relevant variation beyond the label sequence \cite{muller2023considerations}. These constraints have motivated gloss-free approaches that map spoken-language text more directly to sign pose sequences.
Early work in this direction explored transformer-based continuous sign language production \cite{saunders2020progressive,saunders2021continuous,khan2025multigau}.
Beyond spoken-language text, Ham2Pose maps HamNoSys sign notation to pose sequences, demonstrating that structured linguistic inputs can be animated directly into continuous signing motion \cite{arkushin2023ham2pose}.
More recent works tried to improve sentence-level semantic faithfulness via learned motion representations and decoding strategies, including discrete pose codebooks with vector quantization \cite{hwang2025efficient}, non-autoregressive generation with iterative latent refinement \cite{kiziltepe2025iterative}, diffusion-based generation in learned latent spaces \cite{fang2023signdiff,baltatzis2024neural,feng2025text2sign}, and flow-matching-based generation \cite{khan2025signflow}. Overall, these works reduce dependency on gloss annotations and improve sentence-level semantic alignment. However, most methods focus on matching joint configurations and do not explicitly model temporal dynamics, such as repetition or path extent, which can hinder the capture of the diversity of valid realizations in SLG.

\subsection{Guidance and Coupling in Conditional Generation}

Classifier-free guidance (CFG) is widely used to strengthen conditioning in generative models. Popularized in diffusion models, CFG combines unconditional and conditional predictions during sampling with a guidance scale that increases adherence to the conditioning signal while typically reducing diversity at high scales~\cite{ho2022classifier}.
It has since been adopted broadly in conditional generation, including text-to-image and text-to-motion synthesis, because it often improves semantic alignment without modifying the underlying model architecture \cite{saharia2022photorealistic,rombach2022high,ren2023diffusion,tevet2022human}.
More recently, guidance mechanisms have also been studied for flow-based generative models.
Zheng et al.~\cite{zheng2023guided} introduced \emph{guided flows}, framing guidance as an explicit mechanism to steer continuous-time dynamics for both generation and downstream decision-making.
Fan et al.~\cite{fan2025cfg} revisited CFG in the flow-matching setting and proposed variants tailored to flow models, highlighting that guidance behavior in flows can differ from diffusion.
Complementarily, Feng et al.~\cite{feng2025guidance} provided a systematic treatment of guidance in flow matching, unifying multiple guidance formulations and clarifying their connections and trade-offs.
Despite its effectiveness, CFG primarily strengthens conditional adherence at inference, and its success depends on learning compatible unconditional and conditional fields under condition dropout.
In OT-trained flow matching, the training signal is shaped by minibatch couplings; recent work has shown that naive couplings in conditional settings can induce train--test mismatch when the coupling cost ignores structure relevant to the condition \cite{cheng2025curse}.
Our work connects these threads by studying CFG in CFM-OT for SLG and introducing a coupling strategy that incorporates temporal spectral structure to improve the compatibility of conditional and unconditional training signals.

\paragraph{Motivation: Dynamics-aware couplings under CFG training.}
CFG training alternates between unconditional (masked) and conditional (unmasked) updates, which emphasize different motion distributions ($p(x)$ vs.\ $p(x\!\mid\!c)$)~\cite{ho2022classifier}.
With pose-only OT costs, minibatch couplings can mix samples with incompatible dynamics, especially in masked updates, leading to inconsistent pairings across the two training modes.
We therefore propose \textbf{Spectral Optimal Transport (SOT)}, which augments the OT cost with frequency-domain motion features so that couplings respect both pose similarity and temporal dynamics, improving CFG-based generation.

\begin{algorithm}[t]
\caption{Minibatch OT Coupling (Euclidean)}
\label{alg:ot}
\begin{algorithmic}[1]
\Require Source distribution $p_0$, targets $\mathbf{X}_1=\{\mathbf{x}_1^{\,j}\}_{j=1}^B$, conditions $\mathbf{Y}=\{\mathbf{y}^{\,j}\}_{j=1}^B$
\Ensure Coupled pairs $(\mathbf{X}_0, \mathbf{X}_1', \mathbf{Y}')$
\State Sample $\mathbf{X}_0=\{\mathbf{x}_0^{\,i}\}_{i=1}^B \sim p_0$
\State Compute $M_{\text{pose}}(i,j)=\|\mathrm{vec}(\mathbf{x}_0^{\,i})-\mathrm{vec}(\mathbf{x}_1^{\,j})\|_2^2$
\State Solve assignment $\pi \leftarrow \arg\min_{\pi}\sum_{i=1}^B M_{\text{pose}}(i,\pi(i))$ \Comment{Hungarian}
\State Permute: $\mathbf{X}_1'=\{\mathbf{x}_1^{\,\pi(i)}\}_{i=1}^B$, $\mathbf{Y}'=\{\mathbf{y}^{\,\pi(i)}\}_{i=1}^B$
\State \Return $\mathbf{X}_0,\mathbf{X}_1',\mathbf{Y}'$
\end{algorithmic}
\end{algorithm}

\begin{algorithm}[t]
\caption{Spectral OT Coupling (SOT, Ours)}
\label{alg:sot}
\begin{algorithmic}[1]
\Require Targets $\mathbf{X}_1=\{\mathbf{x}_1^{\,j}\}_{j=1}^B$, conditions $\mathbf{Y}=\{\mathbf{y}^{\,j}\}_{j=1}^B$
\Require Source distribution $p_0$, spectral ratio $\rho$, descriptor $\psi(\cdot)$
\Ensure Coupled pairs $(\mathbf{X}_0, \mathbf{X}_1', \mathbf{Y}')$
\State Sample $\mathbf{X}_0=\{\mathbf{x}_0^{\,i}\}_{i=1}^B \sim p_0$
\State Compute $M_{\text{pose}}(i,j)=\|\mathrm{vec}(\mathbf{x}_0^{\,i})-\mathrm{vec}(\mathbf{x}_1^{\,j})\|_2^2$
\State Compute $M_{\text{spec}}(i,j)=\|\psi(\mathbf{x}_0^{\,i})-\psi(\mathbf{x}_1^{\,j})\|_2^2$
\State Set $\lambda_{\text{spec}} \leftarrow \rho \cdot \dfrac{\mathbb{E}[M_{\text{pose}}]}{\mathbb{E}[M_{\text{spec}}]+\epsilon}$
\State Combine $M(i,j) \leftarrow M_{\text{pose}}(i,j) + \lambda_{\text{spec}} M_{\text{spec}}(i,j)$
\State Solve assignment $\pi \leftarrow \arg\min_{\pi} \sum_{i=1}^B M(i,\pi(i))$ \Comment{Hungarian}
\State Permute: $\mathbf{X}_1'=\{\mathbf{x}_1^{\,\pi(i)}\}_{i=1}^B$, $\mathbf{Y}'=\{\mathbf{y}^{\,\pi(i)}\}_{i=1}^B$
\State \Return $\mathbf{X}_0,\mathbf{X}_1',\mathbf{Y}'$
\end{algorithmic}
\end{algorithm}

\section{Method}
\subsection{Preliminaries}
\label{sec:prelim}

\paragraph{Conditional flow matching.}
Flow Matching (FM) and its conditional variant (CFM) learn a time-dependent vector field by regressing to a target velocity along a chosen probability path \cite{lipman2022flow}.
Let $\mathbf{x}_1 \in \mathbb{R}^{d}$ denote a target SL motion sequence (e.g., a $T$-frame pose sequence flattened into a $d$-dimensional vector), and let $\mathbf{y}$ be the text condition.
We denote by $\mathbf{x}(t)\in\mathbb{R}^{d}$ an intermediate point along a chosen probability path that transports samples from a simple source distribution to the data distribution, and by $\mathbf{v}_\theta(t,\mathbf{x}\mid\mathbf{y})\in\mathbb{R}^{d}$ a neural, time-dependent conditional velocity field parameterized by $\theta$.
We draw an initial sample $\mathbf{x}_0 \sim p_0$ from a simple source distribution ($p_0=\mathcal{N}(\mathbf{0},\mathbf{I})$), and generate samples by solving the conditional ODE
\begin{equation}
\frac{d\mathbf{x}(t)}{dt} = \mathbf{v}_\theta\!\bigl(t, \mathbf{x}(t)\mid \mathbf{y}\bigr),
\qquad
\mathbf{x}(0)=\mathbf{x}_0.
\end{equation}

\paragraph{Minibatch optimal transport (OT) coupling.}
To obtain straight probability paths, a minibatch of noise samples is coupled with a minibatch of data samples using OT, replacing random pairing with a cost-minimizing permutation \cite{tong2023improving}.
Given noise $\mathbf{X}_0=\{\mathbf{x}_0^{\,i}\}_{i=1}^{B}$ and data $\mathbf{X}_1=\{\mathbf{x}_1^{\,j}\}_{j=1}^{B}$, OT computes a permutation $\mathbf{P}\in\mathcal{P}_B$ that minimizes a pairwise cost, typically Euclidean distance in pose space.
After coupling, each paired $(\mathbf{x}_0,\mathbf{x}_1)$ defines the linear path
\begin{equation}
\mathbf{x}_t \;=\; \phi_t(\mathbf{x}_0,\mathbf{x}_1)
\;=\; \bigl(1-(1-\sigma_{\min})t\bigr)\mathbf{x}_0 + t\,\mathbf{x}_1,
\end{equation}
with the corresponding ideal (time-invariant) velocity
\begin{equation}
\mathbf{u}_t^{\mathrm{OT}}(\mathbf{x}_t \mid \mathbf{x}_0,\mathbf{x}_1)
\;=\; \mathbf{x}_1 - (1-\sigma_{\min})\mathbf{x}_0.
\end{equation}
CFM then regresses $\mathbf{v}_\theta\!\bigl(t,\mathbf{x}_t \mid \mathbf{y}\bigr)$ to $\mathbf{u}_t^{\mathrm{OT}}$ along these OT-coupled paths.

\paragraph{Classifier-free guidance (CFG) for flow matching.}
We adopt a CFG \cite{ho2022classifier} condition dropout scheme to train a single model that supports both conditional and unconditional vector fields.
Let $b \sim \mathrm{Bernoulli}(p_{\mathrm{uncond}})$ and define the dropped condition
\begin{equation}
\tilde{\mathbf{y}}=
\begin{cases}
\mathbf{y}, & b=0,\\
\emptyset, & b=1,
\end{cases}
\end{equation}
where $\emptyset$ denotes the null condition.
Under the OT-coupled linear path $\mathbf{x}_t=\phi_t(\mathbf{x}_0,\mathbf{x}_1)$,
we optimize
\begin{equation}
\mathcal{L}(\boldsymbol{\theta})
= \mathbb{E}_{t,\,b,\,(\mathbf{x}_1,\mathbf{y})\sim q,\,\mathbf{x}_0\sim p_0}
\Bigl[
\bigl\|
\mathbf{v}_{\boldsymbol{\theta}}\!\bigl(t,\mathbf{x}_t \mid \tilde{\mathbf{y}}\bigr)
- \mathbf{u}_t^{\mathrm{OT}}\!\bigl(\mathbf{x}_t \mid \mathbf{x}_0,\mathbf{x}_1\bigr)
\bigr\|_2^2
\Bigr].
\end{equation}

At inference time, guidance combines unconditional and conditional predictions as
\begin{equation}
\tilde{\mathbf{v}}_{\boldsymbol{\theta}}(t,\mathbf{x}\mid\mathbf{y})
= \mathbf{v}_{\boldsymbol{\theta}}(t,\mathbf{x}\mid\emptyset)
+ \omega\Bigl(\mathbf{v}_{\boldsymbol{\theta}}(t,\mathbf{x}\mid\mathbf{y})
- \mathbf{v}_{\boldsymbol{\theta}}(t,\mathbf{x}\mid\emptyset)\Bigr).
\end{equation}
where $\omega \ge 0$ is the guidance scale (typically $\omega>1$ for stronger conditioning), and we integrate the ODE from $t=0$ to $t=1$ using $\tilde{\mathbf{v}}_{\boldsymbol{\theta}}$.

\subsection{Spectral Optimal Transport Coupling}
\label{sec:sot}
We introduce \textbf{Spectral Optimal Transport (SOT)}, a frequency-aware minibatch OT coupling used during CFM-OT training. SOT modifies the minibatch assignment cost by combining a standard pose-space distance with a temporal spectral distance, so that the coupling accounts for the targets' temporal-frequency structure in addition to pose similarity.
 
\paragraph{Procedure.}
Alg.~\ref{alg:sot} summarizes SOT. For each minibatch $(\mathbf{X}_1,\mathbf{Y})=\{(\mathbf{x}_1^{\,j},\mathbf{y}^{\,j})\}_{j=1}^B \sim q(\mathbf{x}_1,\mathbf{y})$, we (i) sample prior noise $\mathbf{X}_0=\{\mathbf{x}_0^{\,i}\}_{i=1}^B \sim p_0$, (ii) compute a pose-space cost matrix $M_{\text{pose}}$, (iii) compute a temporal spectral descriptor $\psi(\cdot)$ for both $\mathbf{X}_0$ and $\mathbf{X}_1$ to obtain a spectral cost matrix $M_{\text{spec}}$, (iv) scale the spectral term via ratio normalization, and (v) solve a minimum-cost assignment with Hungarian matching \cite{kuhn1955hungarian} to obtain the coupling permutation. The resulting paired samples are then used to construct the CFM-OT training paths and target velocities (Sec.~\ref{sec:prelim}).
 
\paragraph{Temporal DCT descriptor.}
We instantiate $\psi(\mathbf{x})$ with a temporal DCT because it provides a compact, global summary of periodicity and smoothness in motion trajectories, is efficient and hyperparameter-free, and has been widely used in human motion modeling \cite{akhter2008nonrigid, mao2019learning, mao2021generating, xing2024scene,  li2022skeleton}. Given a pose sequence $\mathbf{x}\in\mathbb{R}^{T\times D}$, we compute a DCT-II along time for each feature dimension $d$:
\begin{equation}
\Psi_{k,d}(\mathbf{x})
=
\sqrt{w_k}\,\sum_{t=0}^{T-1} x_{t,d}\,
\cos\!\Bigl(\frac{\pi}{T}\bigl(t+\tfrac{1}{2}\bigr)k\Bigr),
\quad k=0,\dots,T-1,
\label{eq:dctfeat}
\end{equation}
where $w_k = 1 + (\beta-1)\,k/(T-1)$ is a frequency-dependent gain increasing linearly from $1$ at $k=0$ to $\beta$ at $k=T-1$ ($\beta=3$ in our experiments), and we form $\psi(\mathbf{x})$ by concatenating coefficients $\Psi_{k,d}(\mathbf{x})$ across $k$ and $d$. The gain places more weight on faster-varying temporal components, which would otherwise contribute little to the coupling cost since motion energy concentrates at low frequencies.
 
\paragraph{Spectral OT cost.}
Using $\psi(\cdot)$, SOT defines the pose and spectral costs
\begin{align}
M_{\text{pose}}(i,j) &= \left\| \mathrm{vec}(\mathbf{x}_0^{\,i}) - \mathrm{vec}(\mathbf{x}_1^{\,j}) \right\|_2^2,\\
M_{\text{spec}}(i,j) &= \left\| \psi(\mathbf{x}_0^{\,i}) - \psi(\mathbf{x}_1^{\,j}) \right\|_2^2,
\end{align}
and couples minibatches using the combined cost
\begin{equation}
M(i,j)= M_{\text{pose}}(i,j) + \lambda_{\text{spec}}\, M_{\text{spec}}(i,j).
\label{eq:sot_cost}
\end{equation}
 
\paragraph{What the spectral term does to the coupling.}
Instead of comparing the motion dynamics of noise, which is a prior sample, with those of a target, SOT reweights the metric under which the two are coupled: since a minimum-cost assignment is invariant to row- and column-constant terms of the quadratic cost in Eq.~\eqref{eq:sot_cost}, the coupling is determined entirely by the frequency-weighted correlation between the DCT coefficients of $\mathbf{x}_0$ and $\mathbf{x}_1$, with the weight of frequency $k$ set by the gain $w_k$. Under a pose-only cost, all temporal frequencies are weighted equally, so, because sign-motion energy concentrates at low frequencies, the assignment is decided almost entirely by the slow, gross trajectory. The gain $w_k$ raises the share contributed by the components that carry repetition and articulation speed, making targets that differ in motion dynamics more distinguishable during coupling. The motion dynamics thus enter on the \emph{target} side throughout: since $\mathbf{x}_0\sim\mathcal{N}(\mathbf{0},\mathbf{I})$ is isotropic and $\psi$ is linear, $\psi(\mathbf{x}_0)$ acts as a dynamics-neutral probe. In practice, the spectral term is a small perturbation of the pose cost that mainly biases how near-tied assignments are resolved toward pairings consistent with the targets' temporal-frequency structure, which also yields more stable minibatch couplings.

\paragraph{Ratio-normalized spectral weighting.}
Because $M_{\text{pose}}$ and $M_{\text{spec}}$ can differ in scale, we set
\begin{equation}
\lambda_{\text{spec}}
=
\rho \cdot \frac{\mathbb{E}[M_{\text{pose}}]}{\mathbb{E}[M_{\text{spec}}]+\epsilon},
\label{eq:ratio_scaling}
\end{equation}
where $\mathbb{E}[\cdot]$ denotes the minibatch mean, $\rho$ controls the target relative contribution of the spectral term, and $\epsilon$ is a small constant for numerical stability.
 
\paragraph{Coupling and usage.}
We solve the minimum-cost assignment under $M$ (Eq.~\ref{eq:sot_cost}) using Hungarian matching (Alg.~\ref{alg:sot}) to obtain a coupling permutation. The paired $(\mathbf{x}_0,\mathbf{x}_1)$ are then used to construct the CFM-OT training path and target velocity (Sec.~\ref{sec:prelim}).

\section{Experiments}






\begin{table*}[t]
\caption{\textbf{Quantitative results}
DTW-based pose metrics (PA-JPE and JPE)\cite{zuo2025signs}.
BLEU-4 (B-4) represents a semantic metric.
Since pretrained checkpoints are not publicly released for the evaluator and some baselines, we train the required models from scratch.}
\centering
\footnotesize
\setlength{\tabcolsep}{2.5pt}
\renewcommand{\arraystretch}{1.15}
\begin{tabular}{lcccccccccc}
\toprule
\multirow{3}{*}{Method}
& \multicolumn{5}{c}{CSL-Daily}
& \multicolumn{5}{c}{Phoenix-2014T} \\
\cmidrule(lr){2-6}\cmidrule(lr){7-11}
& \multicolumn{2}{c}{PA-JPE$\downarrow$}
& \multicolumn{2}{c}{JPE$\downarrow$}
& B-T$\uparrow$
& \multicolumn{2}{c}{PA-JPE$\downarrow$}
& \multicolumn{2}{c}{JPE$\downarrow$}
& B-T$\uparrow$ \\
\cmidrule(lr){2-3}\cmidrule(lr){4-5}\cmidrule(lr){7-8}\cmidrule(lr){9-10}
& Body & Hand & Body & Hand & B-4
& Body & Hand & Body & Hand & B-4 \\
\midrule
Prog.\ Trans. \cite{saunders2020progressive} & 15.98 & 12.91 & 16.30 & 32.63 & 1.97 & 13.67 & 11.95 & 15.01 & 31.77 & 2.03 \\
Text2Mesh \cite{stoll2022there} & 13.47 & 12.10 & 13.76 & 30.37 & -- & 13.48 & 12.06 & 14.04 & 31.64 & -- \\
T2S-GPT \cite{yin2024t2s} & 11.94 & 5.93  & 12.32 & 15.43 & -- & 10.38 & 6.47  & 11.65 & 19.09 & -- \\
S-MotionGPT \cite{jiang2023motiongpt} & 10.81 & 3.78  & 11.58 & 11.31 & -- & 9.45  & 3.41  & 10.42 & 9.08  & -- \\
SOKE \cite{zuo2025signs} & 6.24 & 1.71 & \underline{7.38} & 9.68 & 6.22 & 4.77 & 1.38 & 6.04 & 7.72 & 6.92 \\
SignFlow \cite{khan2025signflow} & \underline{4.49} & \underline{1.03} & 7.45 & \underline{4.51} & \underline{6.84} & \underline{4.47} & \underline{1.00} & \underline{4.78} & \underline{6.52} & \underline{8.91} \\
\midrule
Ours & \textbf{4.42} & \textbf{1.01} & \textbf{6.87} & \textbf{4.43} & \textbf{9.09} & \textbf{4.32} & \textbf{0.97} & \textbf{4.76} & \textbf{6.06} & \textbf{9.78} \\
\bottomrule
\end{tabular}
\label{tab:sota_comparison_csl_phx}
\end{table*}

\begin{figure*}[t]
    \centering
    \setlength{\tabcolsep}{2pt}

    \begin{subfigure}[t]{\textwidth}
        \centering
        \includegraphics[width=0.77\textwidth]{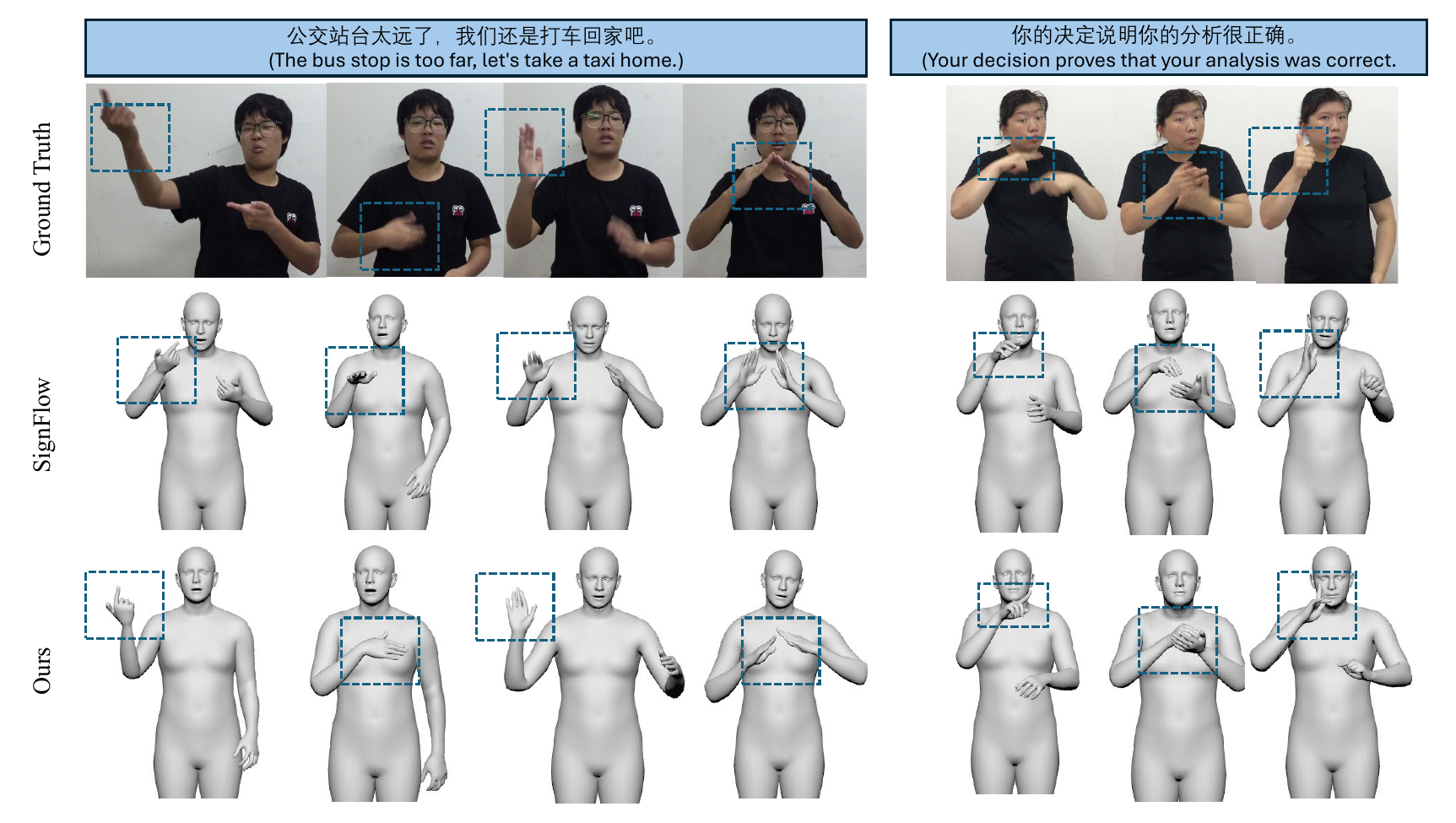}
        \caption{CSL-Daily examples.}
        \label{fig:qual_csl}
    \end{subfigure}


    \begin{subfigure}[t]{\textwidth}
        \centering
        \includegraphics[width=0.77\textwidth]{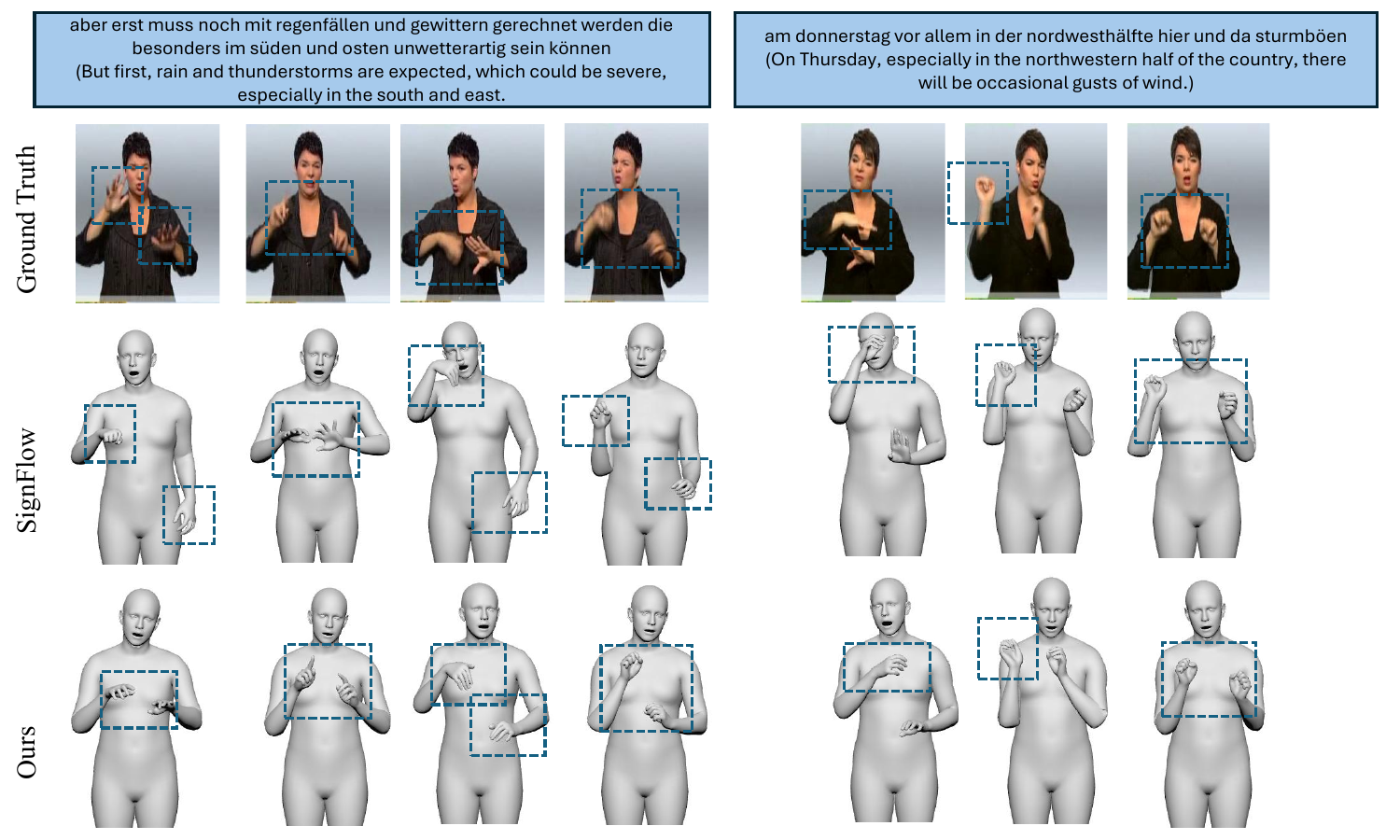}
        \caption{Phoenix-2014T examples.}
        \label{fig:qual_phx}
    \end{subfigure}

    \caption{\textbf{Qualitative comparison.} We show representative examples from CSL-Daily and Phoenix-2014T. }
    \label{fig:qualitative_results}
\end{figure*}

We conduct experiments on two widely used SLG benchmark datasets: CSL-Daily~\cite{zhou2021improving} and Phoenix-2014T~\cite{camgoz2018neural}. 
CSL-Daily is a large-scale Chinese Sign Language dataset of daily conversations with paired spoken-language sentences and sign videos (about 20K samples). 
Phoenix-2014T is a German Sign Language benchmark of weather forecast news (about 8K samples). 
For all experiments, we represent signing as SMPL-X pose sequences. 
We use SMPL-X motion features from the curated preprocessing released by \cite{zuo2025signs}, using 6D rotations and 10 SMPL-X shape parameters.

Following \cite{zuo2025signs}, we evaluate generated sign pose sequences by first aligning each generated sequence to its ground-truth counterpart using Dynamic Time Warping (DTW)~\cite{berndt1994using}. We report DTW-aligned joint position errors under both Procrustes alignment (PA-JPE) and the original joint coordinates (JPE), which quantify sequence-level distances between generated motions and ground truth~\cite{arkushin2023ham2pose}. Following prior SLG work~\cite{baltatzis2024neural}, we evaluate semantic alignment using Back-Translation (BT): a separate sign-to-text model translates each generated pose sequence back into spoken-language text, and we compute ROUGE-L~\cite{lin2004automatic} and BLEU-$n$~\cite{papineni2002bleu} between the reconstructed and reference texts. We train the sign-to-text model of~\cite{tan2025improvement} as our BT evaluator; training details are provided in the supplementary material.

\paragraph{Implementation details.}
We build on the CFM-OT baseline \cite{khan2025signflow}. 
Across all experiments, we keep the architecture, text encoder, training schedule, and optimization settings fixed as mentioned in \cite{khan2025signflow}.
For OT coupling, we assume uniform marginals and compute the minimum-cost bipartite assignment using the Hungarian algorithm \cite{kuhn1955hungarian}. For SOT, we set the pose-space term weight to $1$ and compute the temporal DCT descriptor by controlling the ratio-normalized spectral strength via $\rho$ with $\epsilon=10^{-8}$ (Eq.~\ref{eq:ratio_scaling}). We use $\rho=2\times 10^{-2}$, and we additionally report an ablation over different $\rho$ values in Sec.~\ref{sec:ablation}. For CFG, we train with condition dropout by dropping the text condition with probability $p_{\mathrm{uncond}}=0.1$, and apply guidance at inference with scale $\omega=4$.

\section{Results}

\subsection{Quantitative Results}
\label{sec:quant_results}


Table~\ref{tab:sota_comparison_csl_phx} compares our method with prior approaches on CSL-Daily and Phoenix-2014T. Pose accuracy is measured with the DTW-based metrics of SOKE~\cite{zuo2025signs}, and BLEU-4 (B-4) is computed with a back-translation evaluator trained by us, so that all methods are scored under the same semantic evaluation. Our method obtains the best results on every metric across both datasets. The margin over the strongest baseline, SignFlow~\cite{khan2025signflow}, is modest on
the pose metrics (e.g., PA-JPE Body 4.49$\rightarrow$4.42 on CSL-Daily and 4.47$\rightarrow$4.32 on Phoenix-2014T) but substantial on the semantic one: BLEU-4 improves from 6.84 to 9.09 on CSL-Daily ($+2.25$) and from 8.91 to 9.78 on Phoenix-2014T ($+0.87$). This distribution of gains matches the design of our coupling strategy, which targets semantic alignment under guidance rather than raw joint accuracy: SOT preserves the pose fidelity of the CFM-OT baseline while clearly strengthening sentence-level semantic fidelity.

\begin{table*}[t]
\caption{\textbf{Back-translation results on CSL-Daily and Phoenix-2014T (standard split).}
\textit{Guided} denotes classifier-free guided sampling at inference, while \textit{Unguided} uses the same trained model without guidance ($\omega=0$).
\textit{GT (upper bound)} reports back-translation scores when the evaluator is fed ground-truth motions.
Best and second-best results within each dataset/sampling block (excluding GT) are in \textbf{bold} and \underline{underlined}, respectively.}
\centering
\small
\setlength{\tabcolsep}{5pt}
\renewcommand{\arraystretch}{1.12}
\begin{tabular}{lccccc}
\toprule
\textbf{Method} & \textbf{ROUGE-L} & \textbf{BLEU-1} & \textbf{BLEU-2} & \textbf{BLEU-3} & \textbf{BLEU-4} \\
\midrule

\multicolumn{6}{l}{\textbf{CSL-Daily} \hfill \textit{Unguided sampling}} \\
\midrule
\textit{GT (upper bound)}           & 33.50 & 31.82 & 20.34 & 13.48 & 9.27 \\
CFM-OT \cite{khan2025signflow}      & 26.37 & \textbf{26.39} & \textbf{15.41} & \underline{9.90} & \underline{6.84} \\
CFM-SOT                              & \underline{26.55} & 25.60 & 15.05 & 9.57 & 6.47 \\
CFM-OT + CFG                         & 25.63 & 25.20 & 14.45 & 9.11 & 6.21 \\
CFM-SOT + CFG                        & \textbf{26.79} & \underline{26.09} & \underline{15.30} & \textbf{9.97} & \textbf{6.89} \\
\midrule
\multicolumn{6}{l}{\textbf{CSL-Daily} \hfill \textit{Guided sampling}} \\
\midrule
CFM-OT + CFG                         & \underline{30.46} & \underline{28.89} & \underline{18.09} & \underline{12.10} & \underline{8.44} \\
CFM-SOT + CFG                        & \textbf{31.38} & \textbf{29.48} & \textbf{18.75} & \textbf{12.78} & \textbf{9.09} \\

\midrule\midrule

\multicolumn{6}{l}{\textbf{Phoenix-2014T} \hfill \textit{Unguided sampling}} \\
\midrule
\textit{GT (upper bound)}           & 26.55 & 26.88 & 18.06 & 13.43 & 10.75 \\
CFM-OT \cite{khan2025signflow}      & \textbf{23.43} & \textbf{23.99} & \textbf{15.48} & \underline{11.30} & \underline{8.91} \\
CFM-SOT                              & \underline{23.18} & \underline{23.51} & \underline{15.39} & \textbf{11.40} & \textbf{9.07} \\
CFM-OT + CFG                         & 22.15 & 23.22 & 15.01 & 11.06 & 8.80 \\
CFM-SOT + CFG                        & 22.37 & 22.57 & 14.15 & 10.28 & 8.13 \\
\midrule
\multicolumn{6}{l}{\textbf{Phoenix-2014T} \hfill \textit{Guided sampling}} \\
\midrule
CFM-OT + CFG                         & \underline{23.26} & \underline{23.44} & \underline{15.50} & \underline{11.40} & \underline{8.97} \\
CFM-SOT + CFG                        & \textbf{24.60} & \textbf{24.83} & \textbf{16.40} & \textbf{12.25} & \textbf{9.78} \\
\bottomrule
\end{tabular}
\label{tab:two_datasets_results}
\end{table*}


\begin{figure*}[t]
  \centering
  \begin{subfigure}[t]{0.49\textwidth}
    \centering
    \includegraphics[width=\linewidth]{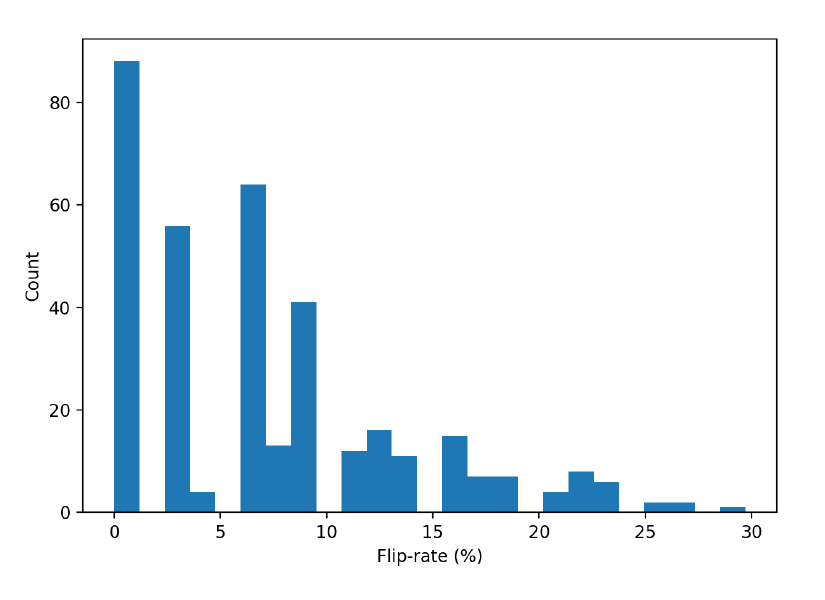}
    \caption{Histogram of coupling flip-rate across minibatches.}
    \label{fig:coupling_hist}
  \end{subfigure}\hfill
  \begin{subfigure}[t]{0.49\textwidth}
    \centering
    \includegraphics[width=\linewidth]{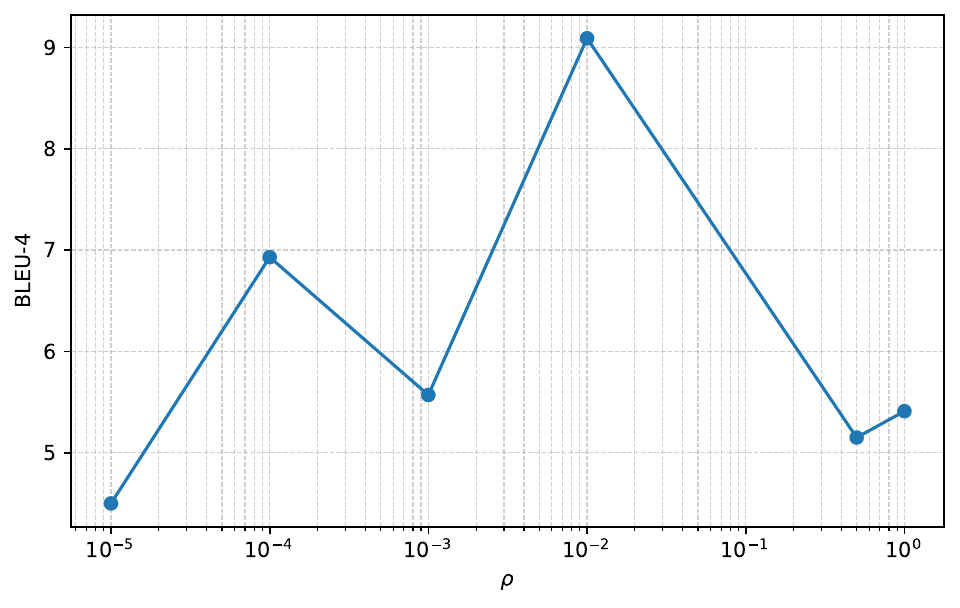}
    \caption{Ablation on the SOT spectral strength $\rho$.}
    \label{fig:ablation_rho}
  \end{subfigure}

  \caption{\textbf{Coupling disagreement and spectral-strength ablation. (on CSL-Daily)}
  \textbf{Left:} Flip-rate is the fraction of rows whose dominant matched partner differs between pose-only OT coupling $P_{\text{pose}}$ and SOT coupling $P_{\text{sot}}$; aggregated over $N{=}300$ minibatches uniformly subsampled during training at $\rho=2\times10^{-2}$.
  \textbf{Right:} BLEU-4 as a function of $\rho$; performance peaks at $\rho=2\times10^{-2}$.
  See the supplementary material for details on the flip-rate computation.}
  \label{fig:flip_and_rho}
\end{figure*}

\subsection{Qualitative Results}
\label{sec:qual}

Figure~\ref{fig:qualitative_results} presents qualitative comparisons on the CSL-Daily and Phoenix-2014T test sets, respectively, showing ground-truth (GT) signer frames alongside avatar renderings generated by SignFlow (CFM-OT) and our method.
Across both datasets, our generations often show clearer signing dynamics: we observe clearer articulation changes over time and more distinct motion patterns in segments where the input sentence implies specific temporal structure (e.g., repetition and path extent).
In addition, our results often produce hand configurations and spatial placements that appear closer to the GT realization, whereas SignFlow tends to yield smoother, more generic trajectories that can blur such dynamics-sensitive variations.
These examples complement the quantitative gains and suggest that incorporating dynamics-aware coupling improves the consistency of conditional generation under guidance. 

\subsection{Ablation Study}
\label{sec:ablation}
\subsubsection{Effect of guidance.}
Table~\ref{tab:two_datasets_results} reports back-translation performance,
which we use throughout this section as a proxy for sentence-level
semantic fidelity.

\paragraph{Guidance helps at inference.}
Applying CFG at inference improves semantic alignment on both datasets.
On CSL-Daily, BLEU-4 rises from 6.84 for unguided CFM-OT to 8.44 for
CFM-OT+CFG with guided sampling, and ROUGE-L from 26.37 to 30.46.
Phoenix-2014T follows the same pattern with a smaller margin: guided
sampling lifts CFM-OT+CFG from 8.80 to 8.97 BLEU-4 and from 22.15 to
23.26 ROUGE-L. Whatever the coupling, guidance makes the generated
sequences adhere more closely to the input text.

\paragraph{Condition dropout hurts training under Euclidean coupling.}
Guidance at inference, however, requires condition dropout at training
time, and this dropout has a cost. To measure it, we evaluate
dropout-trained models with guidance switched off ($\omega=0$). On
CSL-Daily, CFM-OT+CFG evaluated this way falls below the plain CFM-OT
baseline (BLEU-4 6.21 vs.\ 6.84; ROUGE-L 25.63 vs.\ 26.37): sharing
parameters between masked and unmasked updates degrades the model when
the coupling is purely pose-based. Replacing the Euclidean coupling with
SOT recovers the loss, and CFM-SOT+CFG reaches 6.89 BLEU-4 and
26.79 ROUGE-L without guidance, ahead of both the degraded CFM-OT+CFG
and the original baseline. This is what we would expect if SOT reduces
dynamics-inconsistent pairings in masked updates, making the masked and
unmasked training signals more compatible.

\paragraph{SOT and guidance compound.}
The two effects combine at inference. With guided sampling, CFM-SOT+CFG
gives the best results on both datasets: on CSL-Daily it adds +0.65
BLEU-4 (8.44 $\rightarrow$ 9.09) and +0.92 ROUGE-L (30.46
$\rightarrow$ 31.38) over CFM-OT+CFG, and on Phoenix-2014T the margin is
larger still, +0.81 BLEU-4 (8.97 $\rightarrow$ 9.78) and +1.34 ROUGE-L
(23.26 $\rightarrow$ 24.60). Coupling by temporal spectral structure
thus supplies supervision that guidance can exploit at inference, beyond
what either mechanism achieves alone.

\begin{table}[t]
\caption{\textbf{Where to apply SOT under CFG-style masking} (CSL-Daily, \textit{unguided} sampling with $\omega=0$). We apply SOT only to the masked or unmasked coupling step (masking rate 10\%) and compare against unguided baselines.}
\centering
\small
\setlength{\tabcolsep}{6pt}
\renewcommand{\arraystretch}{1.1}
\begin{tabular}{lc}
\toprule
\textbf{Coupling configuration} & \textbf{BLEU-4} \\
\midrule
CFM-OT + CFG (unguided) & 6.21 \\
SOT on masked (10\%) + OT on unmasked & 6.68 \\
SOT on unmasked + OT on masked (10\%) & 6.35 \\
CFM-SOT + CFG (unguided) & \textbf{6.89} \\
\bottomrule
\end{tabular}
\label{tab:masked_unmasked_sot}
\end{table}

\paragraph{Dataset dependence under unguided sampling.}
Phoenix-2014T exhibits a different unguided pattern than CSL-Daily. SOT alone slightly improves BLEU-4 over CFM-OT (8.91 $\rightarrow$ 9.07), while condition-dropout training without guided inference can reduce performance (e.g., CFM-OT+CFG: 8.80; CFM-SOT+CFG: 8.13).
We conjecture that the impact of masking and coupling depends on dataset-specific conditional variability, but the consistent improvements under \emph{guided} inference across both datasets align with our main claim.

\paragraph{Sensitivity to spectral strength $\rho$.}
We ablate the spectral strength $\rho$, which controls the relative contribution of the DCT-based temporal descriptor in the SOT assignment cost via the ratio-normalized weight in Eq.~\eqref{eq:ratio_scaling}.
Intuitively, $\rho$ acts as a knob that determines how strongly the spectral distance influences the OT assignment \emph{relative} to the pose-space distance: larger $\rho$ yields couplings that prioritize spectral similarity more heavily, while smaller $\rho$ makes SOT closer to standard Euclidean OT.
Figure~\ref{fig:ablation_rho} shows BLEU-4 on CSL-Daily as $\rho$ varies.
We observe a clear optimum at $\rho = 2\times 10^{-2}$.
When $\rho$ is increased beyond this range (e.g., $\rho \ge 10^{-1}$), performance degrades, suggesting that over-weighting the spectral term can dominate the coupling and produce assignments that are less compatible with conditional generation.
Conversely, when $\rho$ is too small (e.g., $10^{-5}$), the spectral term becomes weak, and the gain diminishes.

\subsubsection{Evidence that SOT primarily benefits masked-branch coupling.}
To probe which branch benefits most from SOT under CFG-style condition dropout, we apply SOT only when computing the OT assignment for either the masked (unconditional) or unmasked (conditional) branch, while using Euclidean OT for the other branch. We evaluate with \textit{unguided} sampling ($\omega=0$) to isolate training effects from guided inference. Table~\ref{tab:masked_unmasked_sot} shows that applying SOT to the \emph{masked} branch yields a larger improvement over the Euclidean baseline than applying it to the unmasked branch (BLEU-4 6.68 vs.\ 6.35, compared to 6.21 for CFM-OT+CFG). This suggests that coupling errors in masked updates contribute substantially to the degradation observed under shared-parameter dropout training, and that SOT mitigates this by producing more dynamics-consistent assignments. Applying SOT to both branches (CFM-SOT+CFG) achieves the best result (6.89), indicating that both branches benefit, with the masked branch being the dominant contributor in this setting.

\begin{table}[t]
\caption{\textbf{Coupling disagreement induced by SOT (on CSL-Daily).}
Flip-rate measures how often the dominant matched partner changes between $P_{\text{pose}}$ and $P_{\text{sot}}$ (row-wise $\arg\max$ transported mass).
$\Delta_{\text{plan}}$ is the mean absolute difference between the two OT plans, normalized by $B^2$.
Reported as mean$\pm$std over $N{=}300$ logged minibatches uniformly subsampled across training at $\rho=2\times10^{-2}$.}
\centering
\small
\setlength{\tabcolsep}{6pt}
\begin{tabular}{lcc}
\toprule
$\rho$ &
Flip-rate (\%) $\uparrow$ &
$\Delta_{\text{plan}}$ ($\times 10^{-5}$) $\uparrow$ \\
\midrule
$0$ (pose OT)     & $0.00 \pm 0.00$ & $0.00 \pm 0.00$ \\
$2\times 10^{-2}$ & $\mathbf{7.70 \pm 6.70}$ & $\mathbf{3.77 \pm 3.27}$ \\
\bottomrule
\end{tabular}
\label{tab:coupling_disagreement}
\end{table}


\subsubsection{Coupling evidence.}
To verify that the spectral term in SOT meaningfully affects minibatch OT matchings, we measure \emph{coupling disagreement} between pose-only OT and SOT. Concretely, for the same minibatches we compute (i) an OT coupling using only the pose cost and (ii) an OT coupling using the full SOT cost, and quantify their difference using two complementary indicators: (1) the normalized $\ell_1$ distance between the resulting transport plans, and (2) a \emph{flip-rate} measuring how often the dominant matched partner changes. As expected, when $\rho=0$ (pose OT), both metrics are $0$. At our selected operating setting $\rho=2\times 10^{-2}$, we observe a consistent non-zero coupling shift, indicating that SOT alters minibatch pairings and therefore changes the OT-based supervision signal used in conditional flow matching. Figure~\ref{fig:coupling_hist} shows the distribution of flip-rate across $N{=}300$ minibatches uniformly subsampled during training: while some minibatches exhibit zero flips, the histogram contains substantial non-zero mass and a long tail, confirming that SOT can meaningfully modify OT matchings across training. Finally, Figure~\ref{fig:ablation_rho} ablates the spectral strength $\rho$ and reveals a clear optimum at $\rho=2\times 10^{-2}$, consistent with the above evidence that moderate spectral weighting produces beneficial coupling changes. We observe consistent trends on Phoenix-2014T; corresponding ablations are reported in the supplementary material.


\section{Conclusion}

We studied flow-based text-to-sign language generation under a one-to-many setting and investigated how classifier-free guidance (CFG) and minibatch optimal transport (OT) coupling interact during training and sampling. While CFG substantially improves semantic alignment, we hypothesized that standard Euclidean OT coupling can produce dynamics-inconsistent assignments under condition dropout, reducing the compatibility between the conditional and unconditional signals that guidance combines at inference. To address this, we proposed Spectral Optimal Transport (SOT). Across experiments, SOT consistently amplifies the gains of CFG and yields the best quantitative and qualitative results. As future work, we plan to extend the pose-only setting by incorporating facial expressions, which are essential for fully expressive sign language but were outside the scope of this study, which focused on body and hand motion. 

\section*{Acknowledgements}
This work was supported in part by JST Moonshot R\&D under Grant JPMJMS2011, JSPS KAKENHI Grant No. 26H02525, and ROIS NII Open Collaborative Research 2026-261S04-24183.

%
%
\bibliographystyle{splncs04}
\bibliography{main}

@String(AAAI  = {AAAI})

@String(ICASSP=	{ICASSP})

@inproceedings{yin2021including,
  title={Including Signed Languages in Natural Language Processing},
  author={Yin, Kayo and Moryossef, Amit and Hochgesang, Julie and Goldberg, Yoav and Alikhani, Malihe},
  booktitle={Proceedings of the 59th Annual Meeting of the Association for Computational Linguistics and the 11th International Joint Conference on Natural Language Processing (Volume 1: Long Papers)},
  pages={7347--7360},
  year={2021}
}

@article{stokoe2005sign,
  title={Sign language structure: An outline of the visual communication systems of the American deaf},
  author={Stokoe Jr, William C},
  journal={Journal of deaf studies and deaf education},
  volume={10},
  number={1},
  pages={3--37},
  year={2005},
  publisher={Oxford University Press}
}

@article{bosworth2019analysis,
  title={Analysis of the visual spatiotemporal properties of American Sign Language},
  author={Bosworth, Rain G and Wright, Charles E and Dobkins, Karen R},
  journal={Vision research},
  volume={164},
  pages={34--43},
  year={2019},
  publisher={Elsevier}
}

@inproceedings{hwang2025efficient,
  title={An Efficient Gloss-Free Sign Language Translation Using Spatial Configurations and Motion Dynamics with LLMs},
  author={Hwang, Eui Jun and Cho, Sukmin and Lee, Junmyeong and Park, Jong C},
  booktitle={Proceedings of the 2025 Conference of the Nations of the Americas Chapter of the Association for Computational Linguistics: Human Language Technologies (Volume 1: Long Papers)},
  pages={3901--3920},
  year={2025}
}

@article{emmorey1995comparison,
  title={A comparison of spatial language in english \& american sign language},
  author={Emmorey, Karen and Casey, Shannon},
  journal={Sign Language Studies},
  volume={88},
  number={1},
  pages={255--288},
  year={1995},
  publisher={Gallaudet University Press}
}

@inproceedings{li2020word,
  title={Word-level deep sign language recognition from video: A new large-scale dataset and methods comparison},
  author={Li, Dongxu and Rodriguez, Cristian and Yu, Xin and Li, Hongdong},
  booktitle={Proceedings of the IEEE/CVF winter conference on applications of computer vision},
  pages={1459--1469},
  year={2020}
}

@inproceedings{baltatzis2024neural,
  title={Neural Sign Actors: A diffusion model for 3D sign language production from text},
  author={Baltatzis, Vasileios and Potamias, Rolandos Alexandros and Ververas, Evangelos and Sun, Guanxiong and Deng, Jiankang and Zafeiriou, Stefanos},
  booktitle={Proceedings of the IEEE/CVF Conference on Computer Vision and Pattern Recognition},
  pages={1985--1995},
  year={2024}
}

@inproceedings{yin2024t2s,
  title={T2S-GPT: Dynamic Vector Quantization for Autoregressive Sign Language Production from Text},
  author={Yin, Aoxiong and Li, Haoyuan and Shen, Kai and Tang, Siliang and Zhuang, Yueting},
  booktitle={Proceedings of the 62nd Annual Meeting of the Association for Computational Linguistics (Volume 1: Long Papers)},
  pages={3345--3356},
  year={2024}
}

@inproceedings{khan2025multigau,
  title={MultiGAU: Real Time Sign Language Generation Using Multimodal Gated Attention},
  author={Khan, Nabeela and Wu, Bowen and Ishi, Carlos Toshinori and Nakadai, Kazuhiro},
  booktitle={International Conference on Industrial, Engineering and Other Applications of Applied Intelligent Systems},
  pages={149--160},
  year={2025},
  organization={Springer}
}

@inproceedings{qi2024signgen,
  title={Signgen: End-to-end sign language video generation with latent diffusion},
  author={Qi, Fan and Duan, Yu and Zhang, Huaiwen and Xu, Changsheng},
  booktitle={European Conference on Computer Vision},
  pages={252--270},
  year={2024},
  organization={Springer}
}

@inproceedings{khan2025signflow,
  title={SignFlow: End-to-End Sign Language Generation for One-to-Many Modeling using Conditional Flow Matching},
  author={Khan, Nabeela and Wu, Bowen and Tan, Sihan and Ishi, Carlos Toshinori and Nakadai, Kazuhiro},
  booktitle={Proceedings of the 27th International Conference on Multimodal Interaction},
  pages={173--180},
  year={2025}
}

@article{akhter2008nonrigid,
  title={Nonrigid structure from motion in trajectory space},
  author={Akhter, Ijaz and Sheikh, Yaser and Khan, Sohaib and Kanade, Takeo},
  journal={Advances in neural information processing systems},
  volume={21},
  year={2008}
}

@article{tian2024transfusion,
  title={Transfusion: A practical and effective transformer-based diffusion model for 3d human motion prediction},
  author={Tian, Sibo and Zheng, Minghui and Liang, Xiao},
  journal={IEEE Robotics and Automation Letters},
  volume={9},
  number={7},
  pages={6232--6239},
  year={2024},
  publisher={IEEE}
}

@inproceedings{mao2019learning,
  title={Learning trajectory dependencies for human motion prediction},
  author={Mao, Wei and Liu, Miaomiao and Salzmann, Mathieu and Li, Hongdong},
  booktitle={Proceedings of the IEEE/CVF international conference on computer vision},
  pages={9489--9497},
  year={2019}
}

@inproceedings{saunders2020progressive,
  title={Progressive transformers for end-to-end sign language production},
  author={Saunders, Ben and Camgoz, Necati Cihan and Bowden, Richard},
  booktitle={European Conference on Computer Vision},
  pages={687--705},
  year={2020},
  organization={Springer}
}

@inproceedings{xie2024g2p,
  title={G2p-ddm: Generating sign pose sequence from gloss sequence with discrete diffusion model},
  author={Xie, Pan and Zhang, Qipeng and Taiying, Peng and Tang, Hao and Du, Yao and Li, Zexian},
  booktitle={Proceedings of the AAAI Conference on Artificial Intelligence},
  volume={38},
  number={6},
  pages={6234--6242},
  year={2024}
}

@inproceedings{walsh2024select,
  title={Select and Reorder: A Novel Approach for Neural Sign Language Production},
  author={Walsh, Harry and Saunders, Ben and Bowden, Richard},
  booktitle={Proceedings of the 2024 Joint International Conference on Computational Linguistics, Language Resources and Evaluation (LREC-COLING 2024)},
  pages={14531--14542},
  year={2024}
}

@inproceedings{saunders2022signing,
  title={Signing at scale: Learning to co-articulate signs for large-scale photo-realistic sign language production},
  author={Saunders, Ben and Camgoz, Necati Cihan and Bowden, Richard},
  booktitle={Proceedings of the IEEE/CVF Conference on Computer Vision and Pattern Recognition},
  pages={5141--5151},
  year={2022}
}

@inproceedings{tang2022gloss,
  title={Gloss semantic-enhanced network with online back-translation for sign language production},
  author={Tang, Shengeng and Hong, Richang and Guo, Dan and Wang, Meng},
  booktitle={Proceedings of the 30th ACM International Conference on Multimedia},
  pages={5630--5638},
  year={2022}
}

@article{tang2025gloss,
  title={Gloss-driven conditional diffusion models for sign language production},
  author={Tang, Shengeng and Xue, Feng and Wu, Jingjing and Wang, Shuo and Hong, Richang},
  journal={ACM Transactions on Multimedia Computing, Communications and Applications},
  volume={21},
  number={4},
  pages={1--17},
  year={2025},
  publisher={ACM New York, NY}
}

@inproceedings{huang2021towards,
  title={Towards fast and high-quality sign language production},
  author={Huang, Wencan and Pan, Wenwen and Zhao, Zhou and Tian, Qi},
  booktitle={Proceedings of the 29th ACM International Conference on Multimedia},
  pages={3172--3181},
  year={2021}
}

@inproceedings{zuo2024simple,
  title={A simple baseline for spoken language to sign language translation with 3d avatars},
  author={Zuo, Ronglai and Wei, Fangyun and Chen, Zenggui and Mak, Brian and Yang, Jiaolong and Tong, Xin},
  booktitle={European Conference on Computer Vision},
  pages={36--54},
  year={2024},
  organization={Springer}
}

@inproceedings{xie2024sign,
  title={Sign Language Production with Latent Motion Transformer},
  author={Xie, Pan and Peng, Taiying and Du, Yao and Zhang, Qipeng},
  booktitle={2024 IEEE/CVF Winter Conference on Applications of Computer Vision (WACV)},
  pages={3012--3022},
  year={2024},
  organization={IEEE Computer Society}
}

@inproceedings{muller2023considerations,
  title={Considerations for meaningful sign language machine translation based on glosses},
  author={M{\"u}ller, Mathias and Jiang, Zifan and Moryossef, Amit and Gonzales, Annette Rios and Ebling, Sarah},
  booktitle={Proceedings of the 61st Annual Meeting of the Association for Computational Linguistics (Volume 2: Short Papers)},
  pages={682--693},
  year={2023}
}

@inproceedings{kiziltepe2025iterative,
  title={Iterative Latent Refinement for Robust Non-Autoregressive Sign Language Production},
  author={K{\i}z{\i}ltepe, Tu{\u{g}}{\c{c}}e and Ta{\c{s}}y{\"u}rek, S{\"u}meyye Meryem and Keles, Hacer Yalim},
  booktitle={Proceedings of the IEEE/CVF International Conference on Computer Vision},
  pages={4883--4893},
  year={2025}
}

@inproceedings{zhang2025towards,
  title={Towards AI-driven Sign Language Generation with Non-manual Markers},
  author={Zhang, Han and Shalev-Arkushin, Rotem and Baltatzis, Vasileios and Gillis, Connor and Laput, Gierad and Kushalnagar, Raja and Quandt, Lorna C and Findlater, Leah and Bedri, Abdelkareem and Lea, Colin},
  booktitle={Proceedings of the 2025 CHI Conference on Human Factors in Computing Systems},
  pages={1--26},
  year={2025}
}

@inproceedings{arkushin2023ham2pose,
  title={Ham2pose: Animating sign language notation into pose sequences},
  author={Arkushin, Rotem Shalev and Moryossef, Amit and Fried, Ohad},
  booktitle={Proceedings of the IEEE/CVF Conference on Computer Vision and Pattern Recognition},
  pages={21046--21056},
  year={2023}
}

@article{feng2025text2sign,
  title={Text2Sign Diffusion: A Generative Approach for Gloss-Free Sign Language Production},
  author={Feng, Liqian and Wang, Lintao and Hu, Kun and Kong, Dehui and Wang, Zhiyong},
  journal={arXiv preprint arXiv:2509.10845},
  year={2025}
}

@article{fang2023signdiff,
  title={Signdiff: Learning diffusion models for american sign language production},
  author={Fang, Sen and Sui, Chunyu and Zhang, Xuedong and Tian, Yapeng},
  journal={arXiv e-prints},
  pages={arXiv--2308},
  year={2023}
}

@article{saunders2021continuous,
  title={Continuous 3d multi-channel sign language production via progressive transformers and mixture density networks},
  author={Saunders, Ben and Camgoz, Necati Cihan and Bowden, Richard},
  journal={International journal of computer vision},
  volume={129},
  number={7},
  pages={2113--2135},
  year={2021},
  publisher={Springer}
}

@article{ho2022classifier,
  title={Classifier-free diffusion guidance},
  author={Ho, Jonathan and Salimans, Tim},
  journal={arXiv preprint arXiv:2207.12598},
  year={2022}
}

@inproceedings{ren2023diffusion,
  title={Diffusion motion: Generate text-guided 3d human motion by diffusion model},
  author={Ren, Zhiyuan and Pan, Zhihong and Zhou, Xin and Kang, Le},
  booktitle={ICASSP 2023-2023 IEEE International Conference on Acoustics, Speech and Signal Processing (ICASSP)},
  pages={1--5},
  year={2023},
  organization={IEEE}
}

@article{tevet2022human,
  title={Human motion diffusion model},
  author={Tevet, Guy and Raab, Sigal and Gordon, Brian and Shafir, Yonatan and Cohen-Or, Daniel and Bermano, Amit H},
  journal={The Eleventh International Conference on Learning Representations},
  year={2022}
}

@article{fan2025cfg,
  title={Cfg-zero*: Improved classifier-free guidance for flow matching models},
  author={Fan, Weichen and Zheng, Amber Yijia and Yeh, Raymond A and Liu, Ziwei},
  journal={arXiv preprint arXiv:2503.18886},
  year={2025}
}

@article{zheng2023guided,
  title={Guided flows for generative modeling and decision making},
  author={Zheng, Qinqing and Le, Matt and Shaul, Neta and Lipman, Yaron and Grover, Aditya and Chen, Ricky TQ},
  journal={arXiv preprint arXiv:2311.13443},
  year={2023}
}

@article{feng2025guidance,
  title={On the guidance of flow matching},
  author={Feng, Ruiqi and Yu, Chenglei and Deng, Wenhao and Hu, Peiyan and Wu, Tailin},
  journal={arXiv preprint arXiv:2502.02150},
  year={2025}
}

@article{cheng2025curse,
  title={The curse of conditions: Analyzing and improving optimal transport for conditional flow-based generation},
  author={Cheng, Ho Kei and Schwing, Alexander},
  journal={arXiv preprint arXiv:2503.10636},
  year={2025}
}

@article{saharia2022photorealistic,
  title={Photorealistic text-to-image diffusion models with deep language understanding},
  author={Saharia, Chitwan and Chan, William and Saxena, Saurabh and Li, Lala and Whang, Jay and Denton, Emily L and Ghasemipour, Kamyar and Gontijo Lopes, Raphael and Karagol Ayan, Burcu and Salimans, Tim and others},
  journal={Advances in neural information processing systems},
  volume={35},
  pages={36479--36494},
  year={2022}
}

@inproceedings{rombach2022high,
  title={High-resolution image synthesis with latent diffusion models},
  author={Rombach, Robin and Blattmann, Andreas and Lorenz, Dominik and Esser, Patrick and Ommer, Bj{\"o}rn},
  booktitle={Proceedings of the IEEE/CVF conference on computer vision and pattern recognition},
  pages={10684--10695},
  year={2022}
}

@article{lipman2022flow,
  title={Flow matching for generative modeling},
  author={Lipman, Yaron and Chen, Ricky TQ and Ben-Hamu, Heli and Nickel, Maximilian and Le, Matt},
  journal={arXiv preprint arXiv:2210.02747},
  year={2022}
}

@article{tong2023improving,
  title={Improving and generalizing flow-based generative models with minibatch optimal transport},
  author={Tong, Alexander and Fatras, Kilian and Malkin, Nikolay and Huguet, Guillaume and Zhang, Yanlei and Rector-Brooks, Jarrid and Wolf, Guy and Bengio, Yoshua},
  journal={arXiv preprint arXiv:2302.00482},
  year={2023}
}

@inproceedings{zhou2021improving,
  title={Improving sign language translation with monolingual data by sign back-translation},
  author={Zhou, Hao and Zhou, Wengang and Qi, Weizhen and Pu, Junfu and Li, Houqiang},
  booktitle={Proceedings of the IEEE/CVF conference on computer vision and pattern recognition},
  pages={1316--1325},
  year={2021}
}

@inproceedings{camgoz2018neural,
  title={Neural sign language translation},
  author={Camgoz, Necati Cihan and Hadfield, Simon and Koller, Oscar and Ney, Hermann and Bowden, Richard},
  booktitle={Proceedings of the IEEE conference on computer vision and pattern recognition},
  pages={7784--7793},
  year={2018}
}

@inproceedings{zuo2025signs,
  title={Signs as Tokens: A retrieval-enhanced multilingual sign language generator},
  author={Zuo, Ronglai and Potamias, Rolandos Alexandros and Ververas, Evangelos and Deng, Jiankang and Zafeiriou, Stefanos},
  booktitle={Proceedings of the IEEE/CVF International Conference on Computer Vision},
  pages={23806--23816},
  year={2025}
}

@inproceedings{stoll2022there,
  title={There and back again: 3d sign language generation from text using back-translation},
  author={Stoll, Stephanie and Mustafa, Armin and Guillemaut, Jean-Yves},
  booktitle={2022 International Conference on 3D Vision (3DV)},
  pages={187--196},
  year={2022},
  organization={IEEE}
}

@book{sandler2006sign,
  title={Sign language and linguistic universals},
  author={Sandler, Wendy and Lillo-Martin, Diane Carolyn},
  year={2006},
  publisher={Cambridge University Press}
}

@inproceedings{tan2025improvement,
  title={Improvement in sign language translation using text CTC alignment},
  author={Tan, Sihan and Miyazaki, Taro and Khan, Nabeela and Nakadai, Kazuhiro},
  booktitle={Proceedings of the 31st International Conference on Computational Linguistics},
  pages={3255--3266},
  year={2025}
}

@inproceedings{papineni2002bleu,
  title={Bleu: a method for automatic evaluation of machine translation},
  author={Papineni, Kishore and Roukos, Salim and Ward, Todd and Zhu, Wei-Jing},
  booktitle={Proceedings of the 40th annual meeting of the Association for Computational Linguistics},
  pages={311--318},
  year={2002}
}

@inproceedings{lin2004automatic,
  title={Automatic evaluation of machine translation quality using longest common subsequence and skip-bigram statistics},
  author={Lin, Chin-Yew and Och, Franz Josef},
  booktitle={Proceedings of the 42nd annual meeting of the association for computational linguistics (ACL-04)},
  pages={605--612},
  year={2004}
}

@article{kuhn1955hungarian,
  title={The Hungarian method for the assignment problem},
  author={Kuhn, Harold W},
  journal={Naval research logistics quarterly},
  volume={2},
  number={1-2},
  pages={83--97},
  year={1955},
  publisher={Wiley Online Library}
}

@inproceedings{mao2021generating,
  title={Generating smooth pose sequences for diverse human motion prediction},
  author={Mao, Wei and Liu, Miaomiao and Salzmann, Mathieu},
  booktitle={Proceedings of the IEEE/CVF international conference on computer vision},
  pages={13309--13318},
  year={2021}
}

@inproceedings{li2022skeleton,
  title={Skeleton-parted graph scattering networks for 3d human motion prediction},
  author={Li, Maosen and Chen, Siheng and Zhang, Zijing and Xie, Lingxi and Tian, Qi and Zhang, Ya},
  booktitle={European conference on computer vision},
  pages={18--36},
  year={2022},
  organization={Springer}
}

@inproceedings{xing2024scene,
  title={Scene-aware human motion forecasting via mutual distance prediction},
  author={Xing, Chaoyue and Mao, Wei and Liu, Miaomiao},
  booktitle={European Conference on Computer Vision},
  pages={128--144},
  year={2024},
  organization={Springer}
}

@article{jiang2023motiongpt,
  title={Motiongpt: Human motion as a foreign language},
  author={Jiang, Biao and Chen, Xin and Liu, Wen and Yu, Jingyi and Yu, Gang and Chen, Tao},
  journal={Advances in Neural Information Processing Systems},
  volume={36},
  pages={20067--20079},
  year={2023}
}

@inproceedings{berndt1994using,
  title={Using dynamic time warping to find patterns in time series},
  author={Berndt, Donald J and Clifford, James},
  booktitle={Proceedings of the 3rd international conference on knowledge discovery and data mining},
  pages={359--370},
  year={1994}
}
\end{document}